\documentclass[journal,twoside]{IEEEtran}
\usepackage{amsthm,amssymb,mathtools}
\usepackage{subfigure}
\usepackage{graphicx}
\usepackage{subfig}
\usepackage{amsmath}
\usepackage{array}
\usepackage{color}
\newcolumntype{L}[1]{>{\raggedright\let\newline\\\arraybackslash\hspace{0pt}}m{#1}}
\newcolumntype{C}[1]{>{\centering\let\newline  \\\arraybackslash\hspace{0pt}}m{#1}}
\newcolumntype{R}[1]{>{\raggedleft\let\newline \\\arraybackslash\hspace{0pt}}m{#1}}

\begin{document}
\title{Multi-temporal Sentinel-1 and -2 Data Fusion for Optical Image Simulation}
\author{Wei~He,~\IEEEmembership{Member,~IEEE,}
        Naoto~Yokoya,~\IEEEmembership{Member,~IEEE}

\thanks{Manuscript received XX XX, XX; revised XX XX, XX and XX XX, XX, accepted XX XX, XX. This work was supported by the project XXX. (Corresponding author: XX XX.)}
\thanks{W. He and N. Yokoya were with the RIKEN Center for Advanced Intelligence Project, RIKEN, 103-0027 Tokyo, Japan (e-mail:wei.he@riken.jp; naoto.yokoya@riken.jp).\par
Color versions of one or more of the figures in this manuscript are available online at http://ieeexplore.ieee.org. Digital Object Identifier XXXX
}
}
\markboth{~Vol.~XX, No.~X, XXXX~2018}%
{He and Yokoya: Multi-temporal Sentinel-1 and -2 Data Fusion for Optical Simulation}
\maketitle

\begin{abstract}
In this paper, we present the optical image simulation from a synthetic aperture radar (SAR) data using deep learning based methods. Two models, \MakeLowercase{\textit{i.e.}}, optical image simulation directly from the SAR data and from multi-temporal SAR-optical data, are proposed to testify the possibilities. The deep learning based methods that we chose to achieve the models are a convolutional neural network (CNN) with a residual architecture and a conditional generative adversarial network (cGAN). We validate our models using the Sentinel-1 and -2 datasets. The experiments demonstrate that the model with multi-temporal SAR-optical data can successfully simulate the optical image, meanwhile, the model with simple SAR data as input failed. The optical image simulation results indicate the possibility of SAR-optical information blending for the subsequent applications such as large-scale cloud removal, and optical data temporal super-resolution. We also investigate the sensitivity of the proposed models against the training samples, and reveal possible future directions.

\end{abstract}

\begin{IEEEkeywords}
Sentinel, synthetic aperture radar, optical, data simulation, convolutional neural network, generative adversarial network.
\end{IEEEkeywords}

\section{Introduction}
\IEEEPARstart{T}{he} optical data provided by Sentinel-2 has 13 spectral bands from visible, near infrared to short wave infrared spectrum, with a 5-day revisit time at the equator~\cite{RSE_Sentinel2}. Sentinel-2 is useful in time-series analysis such as land cover changes and damage area detection. Change analysis using optical data assumes that all investigated images are cloud-free to classify every pixel in the image, which is often not possible, especially for the cloudy areas of the earth. Usually, there is only one low-cloudy image nearly every month in the cloudy area. Some researchers have reportedly used data from alternative months (previous or next) to composite the data corrupted by clouds~\cite{RSE2010cloud,ISPRS2014shen}. However, these methods remove only small clouds and also ignore the changes between monthly data. In order to overcome these serious limitations, it is necessary to combine other remote sensing data resources, and conduct multi-source data fusion to predict cloud-free Sentinel-2 images.

The last few decades has witnessed a rapid growth in SAR data. SAR data captured by Sentinel-1 exhibits totally different characteristics from that of the optical data. Sentinel-1 has the ability to provide routine, day and night, all-weather resolution observation, and can also overcome various kinds of bad weather conditions such as clouds, rain, smoke and fog~\cite{RSE2012SAR}. In particular, it is expected to provide near daily coverage over Europe and Canada~\cite{RSE2012SAR}. Therefore, one obvious question arises: \textit{can we use SAR data to predict the optical image?}

Recently, many researchers have contributed to the information fusion of SAR and optical images with different motives. Researchers~\cite{RSE2012SAR} have adopted Intensity Hue Saturation (IHS) to integrate hyperspectral, and Topographic SAR into a single image to enhance urban surface features. Some groups~\cite{GRSL2015OPTSAR} tried removal of the speckle noise from SAR data via fusion of two data sources. Reportedly \cite{JSTAR2018SAROPT} the SAR and optical data can be matched by deep learning methods to generate SAR-like image, for the generation of precise ground control points.

Many researchers conduct the fusion of SAR and optical data to produce middle image~\cite{JSTAR2018SAROPT} or final application results. Whether or not the SAR data can be directly translated to optical data still remains a concern. In this work, we aim to investigate the possibility of optical data simulation from SAR data. We have adopted deep learning based methods for the optical simulation for the following three reasons. First, a deep CNN can efficiently capture the image characteristics. Second, several smart techniques have been proposed for training CNN such as, batch normalization (BN)~\cite{ICML2015BN}, residual networks (ResNets)~\cite{He2016Resnets} and Rectifier Linear Unit (ReLU)~\cite{NIPS2012CNN}; recently, a generative adversarial network (GAN) has been proposed and demonstrated as useful in data generation. Third, a deep architecture can be accelerated by graphics processing unit (GPU).

We chose to use two methods, CNN with ResNets and cGAN ~\cite{Isola_2017_CVPR} to complete the task. Equipped with state-of-the-art data simulation algorithms we investigated the possibility of optical image simulation from single SAR imagery acquired at a similar period, and multi-temporal SAR-optical images (SAR imagery with the side information from previous or next pairs of SAR and optical images). Our experiments on Sentinel-1 and -2 data demonstrate the necessity of using multi-temporal images as input and the effectiveness of cGAN.

\section{Approaches}
\subsection{Problem Formulation}

\begin{figure}[!ht]
  \centering
  \includegraphics[width= 0.9 \linewidth]{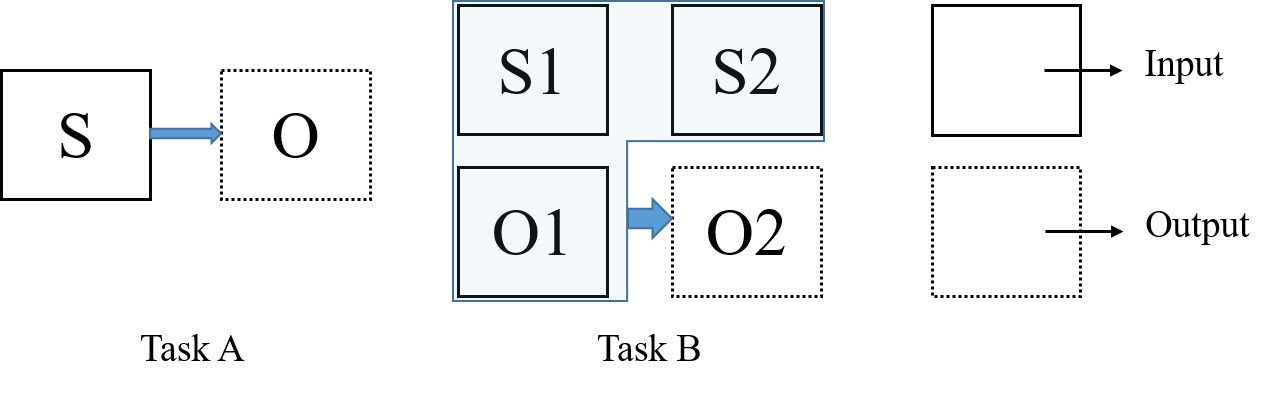}
  \caption{Illustration of two optical simulation tasks}
  \label{fig:Task1}
\end{figure}
The purpose of this work is to simulate an optical image using either a single SAR image or multi-temporal SAR-optical images, which is outlined in Fig.\ref{fig:Task1}. The figure illustrates two tasks of optical image simulation. Task A, shows the optical image simulation directly from the SAR data, and Task B, displays the simulation from SAR (S2) combined with the additional information from the previous pairs of SAR and optical data (S1 and O1). Task B is also referred to as multi-temporal fusion based optical image simulation. The CNN and cGAN are adopted to complete the simulation tasks, and the details of the investigated methods are presented in the subsequent section.\par

\subsection{cGAN}
Conditional GAN is extended from GAN~\cite{NIPS2014GAN} and deep convolutional GAN (DCGAN)~\cite{ICLR2015DCGAN}, which describes a mini-max game between a generative model $G$ and a discriminative model $D$. The generator $G$ is trained from the input image $x$, and random noise $z$ to generate the output image $y$: $G: {x,z} \rightarrow y$. The discriminator $D$ is trained to distinguish the fake image $G(x,z)$ from the real image y. The adversarial processing of the cGAN is presented as follows. The discriminator $D$ tries to distinguish the realistic input-real pairs as 1, i.e., $D(x,y)=1$, and detect the simulated input-fake pairs as 0, i.e., $D(x,G(x,z))=0$. From a second prospective, the generator $G$ tries to generate a fake image to fool the discriminator $D$, in order to increase the accuracy of $D(x,G(x,z))$ to 1. If at any instance the discriminator $D$ cannot distinguish between input-real and input-fake pairs, then the fake image generated by $G$ can be regarded as the predicted optical image (we call it real image). The cGAN loss of this adversarial processing can be detailed as:
\begin{multline}
\min \limits_{G} \max \limits_{D} \mathcal{L}_{cGAN}(G,D) = \bold{E}_{(x,y) \in {p_{data}(x,y)}}[logD(x,y)] \\
+ \bold{E}_{x\in {p_{data}(x)},z\in {p_{data}(z)}}[1-logD(x,G(x,z))]
\end{multline}
Here, $log$ function is adopted to relax the gradient insufficient at the beginning of the training~\cite{NIPS2014GAN}. From a second perspective, the generator's objectives are not only to fool the discriminator, but also to generate the image near the real output $y$ in the sense of $L1/L2$ distance. To encourage less blurring, $L1$ distance, is absorbed into the cGAN loss,
\begin{multline}
\mathcal{L}_{L1}(G) = \bold{E}_{(x,y) \in {p_{data}(x,y)},z\in {p_{data}(z)}}\Vert y-G(x,z)\Vert _{L1}
\end{multline}
resulting in the final objective function:
\begin{equation}\label{finalloss}
G^* = \min \limits_{G} \max \limits_{D}{\mathcal{L}_{cGAN}(G,D)+\lambda \mathcal{L}_{L1}(G)}.
\end{equation}
Here, parameter $\lambda$ demonstrates the trade-off between the cGAN loss, and $L1 $ loss.\par

\subsection{Network Architectures}
This work requires simultaneous training of the generative and discriminative networks. As stated in the preceding section, the generator produces a fake image from the input data, and the discriminator tries to classify the input-fake pair and input-real pair. The discriminator is first trained to improve the classification accuracy. A trained discriminator is then used to help to train the generative network. The process alternates until the end.
Main architectures of generator and discriminator are ~\cite{ICLR2015DCGAN} with the modules of the form Conv(convolution)-BN-ReLu. To keep the spatial size of input and output images, pooling step is left out and stride size is set as 1. We used zero-padding to make up for the spatial size reduction cased by the convolution kernel. \par
\begin{figure}
  \centering
  \includegraphics[width= 0.9 \linewidth]{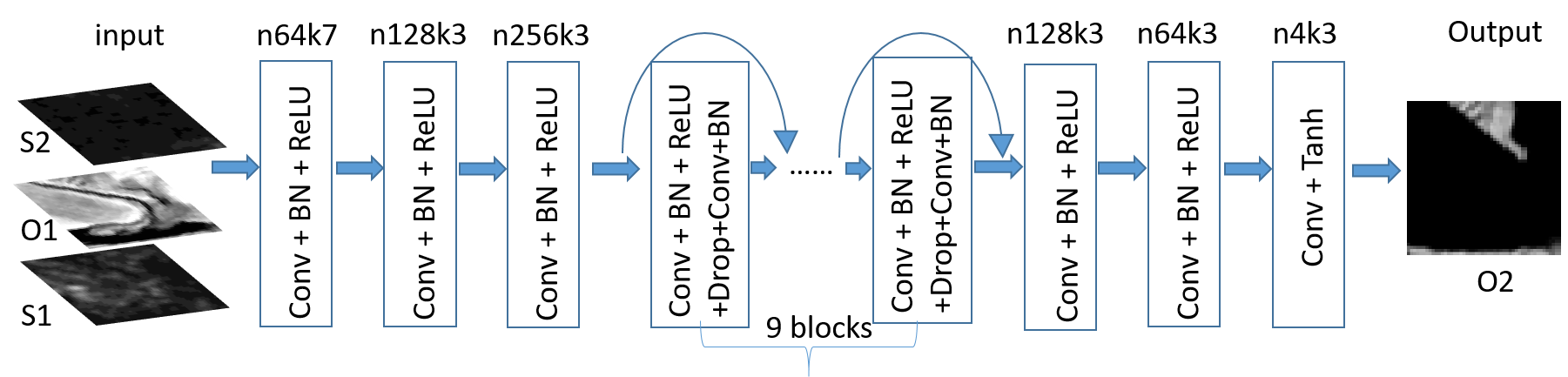}
  \caption{Illustration of the CNN generation network}
  \label{fig:Gnet}
\end{figure}

The discriminative network used is the same as the one introduced in~\cite{Isola_2017_CVPR} with $patchGAN$ to capture high-frequencies and reduced parameters. The generative network is illustrated in Fig.\ref{fig:Gnet}. In this figure, n64k7 means the corresponding number of output features is 64 and the kernel size is 7. The input image is of size $256 \times 256 \times n$ (e.g., $n = 2$ for Task A when two polarimetric channels are used; $n = 8$ for Task B when four polarimetric channels and four spectral bands are used), and the output image is of size $256 \times 256 \times 4$. ResNets have been demonstrated as very useful in the restoration task~\cite{TIP2017DnCNN}. However, ResNets identify network by shortcut, which is inconsistent with our generator network (the features of input is not equal to that of output). In the first three layers, the features rise to $256$ dimensions, followed by $9$ ResNets blocks. Each ResNets block layer is completed by the modules of the form Conv-BN-ReLu-Drop-Conv-ReLu ~\cite{ECCV2016Lifeifei}. In this case, the ResNets are used in the $256$ feature space, and concluded by three layers to reduce the feature dimension to $4$. In particular, Tanh function is adopted instead of ReLu at the output layer as reported in~\cite{ICLR2015DCGAN}. \par

\begin{table}
\centering
\caption{}{Sensing Time of Optical and SAR Image Pairs Used in the Experiments}
\label{tab:Sensing_time}
\begin{tabular}{|c| c | c | c |c| }                                              \hline
	Y-M-D        & S1          & O1           & S2           & O2           \\ \hline\hline
	Iraq         & 2017-11-12  & 2017-11-10   & 2017-12-06   & 2017-12-10   \\
	Jianghan     & 2017-11-14  & 2017-11-12   & 2017-12-20   & 2017-12-19   \\
	Xiangyang    & 2017-11-14  & 2017-11-12   & 2017-12-20   & 2017-12-19   \\ \hline
\end{tabular}
\end{table}
\subsection{Implementation Issues}
Previous works on GAN have demonstrated the importance of using Gaussian noise as input in the generative network. In this work on cGAN, the input is $x$ and the noise is absorbed into the dropout part, which can also produce reasonable results~\cite{Isola_2017_CVPR}. In our experiments, the dropout rate is set as $0.5$. Mini-batch stochastic gradient decent with Adam solver is adopted to train the particular model. The model is trained on $200$ epochs with batch size $1$ and learning rate $0.0002$. As suggested in literature~\cite{dlr119293}, $\lambda$ in the loss objective (\ref{finalloss}) is set to 100 to encourage both reconstruction accuracy and object sharpness, simultaneously.

\section{Dataset Description and Experimental Setting}
\subsection{Dataset}
\begin{figure*}
  \centering
  \includegraphics[width= 1 \linewidth]{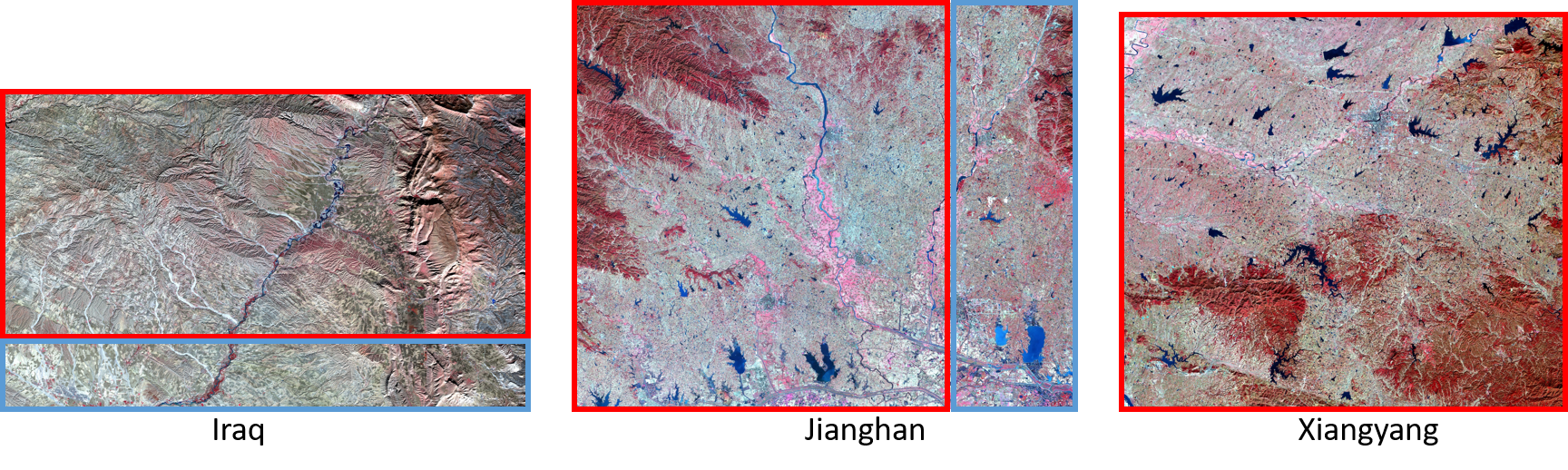}
  \caption{O2 images of Iraq, Jianghan and Xiangyang pairs. The training patches are selected from the red rectangle and the test patches are from the blue area}
  \label{fig:dataintroduction}
\end{figure*}

Sentinel-1 and -2 data\footnote{download website:https://scihub.copernicus.eu/dhus/}
were adopted in our experiments to confirm the possibility of optical simulation from SAR data. The data were pre-processed and co-registered by Sentinel Application Platform (SNAP) software provided by ESA ~\cite{zuhlke2015snap}. We processed the SAR image with the flowchart of calibration-despeckling-Range Doppler Terrain, and two bands of VV/VH intensities with a pixel spacing of $10m$. For Sentinel-2 data, we chose $4$ bands (R-G-B-NIR) with a ground sampling distance of $10m$ for the experiments. The SAR and optical images were co-registered by reprojection; SAR and Optical data pairs from three areas (Iraq, Jianghan, and Xiangyang) were used in the experiments. The acquisition time for each image is presented in Table \ref{tab:Sensing_time}. The absolute difference in acquisition time between S1 and O1 (or, S2 and O2) is ensured to be less then 5 days. Images from Iraq, Jianghan and Xiangyang are of size $8460 \times 5121$, $10657 \times 8659$ and $6801 \times 7651$, respectively. An earthquake happened in Iraq area between time T1 and T2, that caused many changes in the terrain; images from Jianghan and Xiangyang, two similar areas of China were sensed simultaneously. O2 images of data pairs from each of these areas are presented in Fig.\ref{fig:dataintroduction}.

\subsection{Training and Test Setup}

\begin{table}
\centering
\caption{}{The Training and Test Patches provided by the Images}
\label{tab:Training_number}
\begin{tabular}{|c| c | c | c | }                                       \hline
	             & Iraq          & Jianghan           & Xiangyang    \\ \hline\hline
	Train        & 561           & 1188               & 754          \\
	Test         & 99            & 165                & None         \\ \hline
\end{tabular}
\end{table}

The images were segmented into non-overlapping patch pairs of spatial size $256 \times 256$. The training data were then selected from the area inside the red rectangle, and the test data from the blue ones. The number of training, and test patches are summarized in Table \ref{tab:Training_number}. Models specific to each test area were designed according to the respective training dataset as per the details given below. \par
Case 1) The test patches were taken from Iraq image. The optical image simulation results of Tasks A and B were verified with different models, \MakeLowercase{\textit{i.e.}}, CNN (the generation model described in Fig.\ref{fig:Gnet}) and cGAN. For Task A, the training patch pairs were from the whole Iraq image pairs of T1 and areas of T2 marked with the red rectangle, with $1221$ pairs in total. For Task B, the training pairs were only from the areas of Iraq image marked with the red rectangle. The four methods were denoted as CNN (Task A with CNN), cGAN (Task A with cGAN), MTCNN (Task B with CNN), and MTcGAN (Task B with cGAN), respectively.\par
Case 2) The test patches were taken from Jianghan image. In this case, we performed only Task B, and train MTCNN and MTcGAN models with four different training sets. For the first three sets, the samples were selected from the training parts of the Jianghan, Iraq, and Xiangyang images, respectively. In this case, the simulated optical images of MTCNN and MTcGAN methods can be with different training sets. We also added the whole training patches together to formulate the final training set. The experimental studies conducted are thus expected to testify the influence of different training sets for the final optical image simulation.

\subsection{Evaluation Index}
\begin{table}
\centering
\caption{}{Simulation Accuracy of Different Methods in Case 1}
\label{tab:Case1_index}
\begin{tabular}{|c| c | c | c |c|| c|}                                        \hline
	Index     & CNN    & cGAN   & MTCNN   &MTcGAN           & O1           \\ \hline\hline
	PSNR(dB)  & 26.60  & 26.79  & 30.61   &\textbf{32.32}   & 29.77        \\
	SSIM      & 0.6477 & 0.6519 & 0.9028  &\textbf{0.9110}  & 0.8528       \\
	MSA       & 0.6769 & 0.6581 & 0.3796  &\textbf{0.3146}  & 0.5529       \\ \hline
\end{tabular}
\end{table}

\begin{figure*}
  \centering
  \includegraphics[width= 1 \linewidth]{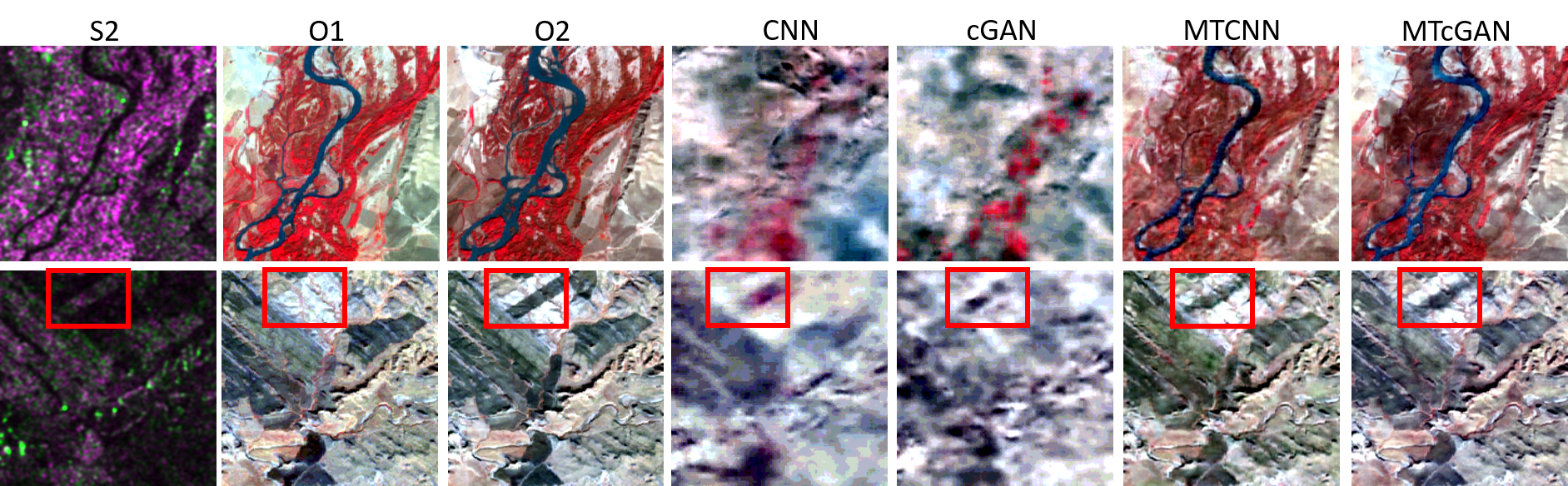}
  \caption{Simulated images of different methods in Case 1, companied with the input images (S1,S2 and O1) and output reference image (O2)}
  \label{fig:Iraqresults}
\end{figure*}

\begin{table}
\centering
\scriptsize
\caption{}{Simulation Accuracy of MTCNN and MTcGAN with Different Training Samples in Case 2
}
\label{tab:Case2_index}
\begin{tabular}{|c| c | c | c |c| c|| c|}                                      \hline
	Method  &Index & Jianghan        & Iraq    & Xiangyang &mixed    &O1    \\ \hline\hline
	        &PSNR  &\textbf{35.08}   & 29.44   & 34.30     & 34.38   &34.01  \\
	MTCNN   &SSIM  &\textbf{0.9508}  & 0.8585  & 0.9412    & 0.9479  &0.9401 \\
	        &MSA   &\textbf{0.4684}  & 0.8400  & 0.5138    & 0.4774  &0.5319 \\ \hline

	        &PSNR  &\textbf{35.25}   & 31.09   & 34.44     & 34.83   &34.01   \\
	MTcGAN  &SSIM  &\textbf{0.9509}  & 0.8850  & 0.9413    & 0.9463  &0.9401  \\
	        &MSA   &\textbf{0.4629}  & 0.6137  & 0.5070    & 0.4649  &0.5319   \\ \hline
\end{tabular}
\end{table}

In this paper, three evaluation indicators: the peak signal-to-noise ratio (PSNR), the structural similarity (SSIM), and the mean spectral angle (MSA), were used to access the quality of the simulated optical image. For the multispectral image, we calculated the values of PNSR and SSIM of each band between simulated optical image, and the reference image, and determined the average~\cite{He2016TGRS}.

\section{Results}
The training program was completed on a single GTX1080 GPU. The PSNR, SSIM and MSA values of different simulation results for Case 1, are evaluated and listed in Table \ref{tab:Case1_index}. The values of the three indices for input O1 data are regarded as baseline for the other sets. The best of the values for each quality index in the table is shown in bold. Table \ref{tab:Case1_index}, also shows that CNN and cGAN achieve lower values for all three quality indices compared to the baseline. That essentially concludes that Task A, which describes the optical image simulation from single SAR imagery fails to predict the image. On the other hand Task B related methods, i.e., MTCNN and MTcGAN are found to achieve higher values for each index type compared to the baseline. The results indicate that compared to the input images, MTCNN and MTcGAN can successfully simulate the optical images. Furthermore, higher index value of MTcGAN than that of MTCNN, suggests the advantage of adversarial network in our simulation task.

Fig. \ref{fig:Iraqresults} shows several patches of input S2 and O1, and output reference O2, compared with our optical image simulation results. In Fig. \ref{fig:Iraqresults}, MTCNN and MTcGAN demonstrate much better results than that of CNN and cGAN from a visual perspective. In fact, SAR image and optical image being totally different from each other, it is extremely difficult to learn a mapping between the two. The optical simulation results of Task A presented in Fig. \ref{fig:Iraqresults} are hence blurred, and one can not distinguish objects from these simulated image. However, with multi-temporal fusion based optical simulation of Task B, one can learn the changed information between S1 and S2, and then accordingly reconstruct the change on the basis of O1 image. As illustrated in the red rectangle of Fig. \ref{fig:Iraqresults}, one can see a change has happened between O1 and O2. Our goal is to pass this change from the SAR image to the optical image, and then reconstruct the same in the simulated O2 image. Following this strategy, the complexity of Task B can be significantly reduced, and the optical image can thus be successfully simulated. \par

\begin{figure*}
  \centering
  \includegraphics[width= 1 \linewidth]{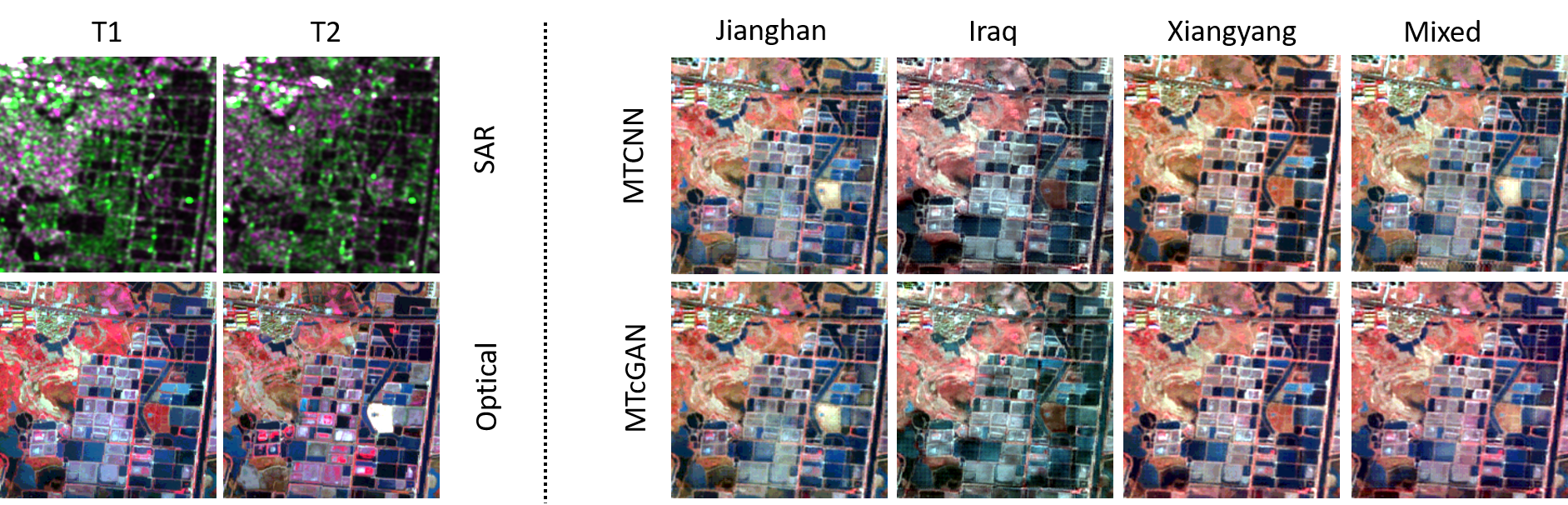}
  \caption{Simulated images of different methods in Case 2. The input images (S1,S2 and O1) and output reference image (O2) at left side, and simulated images with different training samples at right side.}
  \label{fig:Jianghanresults}
\end{figure*}

We also investigated the influence of different training samples on the final optical image simulation results of MTCNN and MTcGAN methods in Case 2. Table \ref{tab:Case2_index} presents the values for the three quality indices, and Fig. \ref{fig:Jianghanresults} illustrates the simulated optical images of different methods from different training sets. It can be easily seen that the two methods with Jianghan image as training samples result in the best values and visual quality. On the other hand, MTCNN and MTcGAN methods with Iraq image as training samples have the lowest PSNR, SSIM and MSA values. Interestingly, the simulation results with Iraq training samples are of high visual quality, but with a significant change in spectral information compared to the reference optical image. This model has thus simulated a new style of optical image, guided by the Iraq training set. The main reason behind the observed result is that the test samples are composed of flat areas, even though, the Iraq training samples are filled with mountains. Xiangyang training samples are more similar to the test Jianghan image patches. As a result, the models with Xiangyang training samples can produce much better result than that with the Iraq samples. Thus, selection of training samples can largely influence the final simulation results; more similarity between test data of the training samples result in better optical image simulation. Unfortunately, accumulation of training sets fails to solve the problem, which is illustrated by the fact that the simulation results obtained with the whole training data together is worse than the one with only Jianghan data.

The experimental part is thus concluded with the verification of two hypotheses. First, a multi-temporal fusion based optical simulation in Task B is valid and effective. Second, GAN based method can produce better results than that of CNN. However, the corresponding simulation results are not so perfect. As illustrated in the red rectangle of Fig. \ref{fig:Iraqresults}, MTcGAN, standing for the best method, can only simulate a blurred object of the change information compared to the reference one. Additionally, the model is sensitive to the training samples; if the training samples are improper, it may lead to production of some fake results with the trained model.

\section{Conclusion}
We have thus investigated the possibility of optical image simulation from single SAR imagery and multi-temporal SAR-optical images, in this paper. Two deep learning based methods have been designed for the said tasks, \MakeLowercase{\textit{i.e.}}, CNN with ResNets and cGAN. We tested our models on Sentinel-1 and -2 datasets and drew the following conclusions. First, multi-temporal data fusion based optical image simulation can successfully generate the optical images. The simulated optical images obtained show more similarity to the reference optical images, both in visual and quantitative evaluation, compared to the input SAR and optical images. Second, an adversarial network is proved useful and effective in our task.

Despite the satisfactory performance of multi-temporal fusion model with the cGAN method, there is still much room for improvement. The simulated optical images especially in the changing part of S1 and S2 images, are blurred and need improvement. Selection of the training samples is also a big concern for our model since without proper samples, the models may create fake optical images. Finally, in our model, we have chosen only two time period information, \MakeLowercase{\textit{i.e.}}, T1 and T2, and it may be possible to choose a few more to obtain better simulation results.


\bibliographystyle{IEEEtran}
\bibliography{lowrank_review_references}

\begin{thebibliography}{10}
\providecommand{\url}[1]{#1}
\csname url@samestyle\endcsname
\providecommand{\newblock}{\relax}
\providecommand{\bibinfo}[2]{#2}
\providecommand{\BIBentrySTDinterwordspacing}{\spaceskip=0pt\relax}
\providecommand{\BIBentryALTinterwordstretchfactor}{4}
\providecommand{\BIBentryALTinterwordspacing}{\spaceskip=\fontdimen2\font plus
\BIBentryALTinterwordstretchfactor\fontdimen3\font minus
  \fontdimen4\font\relax}
\providecommand{\BIBforeignlanguage}[2]{{%
\expandafter\ifx\csname l@#1\endcsname\relax
\typeout{** WARNING: IEEEtran.bst: No hyphenation pattern has been}%
\typeout{** loaded for the language `#1'. Using the pattern for}%
\typeout{** the default language instead.}%
\else
\language=\csname l@#1\endcsname
\fi
#2}}
\providecommand{\BIBdecl}{\relax}
\BIBdecl

\bibitem{RSE_Sentinel2}
M.~Drusch, U.~Del~Bello, S.~Carlier, O.~Colin, V.~Fernandez, F.~Gascon,
  B.~Hoersch, C.~Isola, P.~Laberinti, P.~Martimort, A.~Meygret, F.~Spoto,
  O.~Sy, F.~Marchese, and P.~Bargellini, ``Sentinel-2: Esa's optical
  high-resolution mission for gmes operational services,'' \emph{Remote Sens.
  Environ.}, vol. 120, pp. 25--36, Feb. 2012.

\bibitem{RSE2010cloud}
O.~Hagolle, M.~Huc, D.~V. Pascual, and G.~Dedieu, ``A multi-temporal method for
  cloud detection, applied to formosat-2, ven¦Ìs, landsat and sentinel-2
  images,'' \emph{Remote Sens. Environ.}, vol. 114, no.~8, pp. 1747--1755, Aug.
  2010.

\bibitem{ISPRS2014shen}
Q.~Cheng, H.~Shen, L.~Zhang, Q.~Yuan, and C.~Zeng, ``Cloud removal for remotely
  sensed images by similar pixel replacement guided with a spatio-temporal mrf
  model,'' \emph{ISPRS Journal of Photogrammetry and Remote Sensing}, vol.~92,
  pp. 54--68, Jun. 2014.

\bibitem{RSE2012SAR}
R.~Torres, P.~Snoeij, D.~Geudtner, D.~Bibby, M.~Davidson, E.~Attema, P.~Potin,
  B.~Rommen, N.~Floury, M.~Brown, I.~N. Traver, P.~Deghaye, B.~Duesmann,
  B.~Rosich, N.~Miranda, C.~Bruno, M.~L'Abbate, R.~Croci, A.~Pietropaolo,
  M.~Huchler, and F.~Rostan, ``Gmes sentinel-1 mission,'' \emph{Remote Sens.
  Environ.}, vol. 120, pp. 9--24, May 2012.

\bibitem{GRSL2015OPTSAR}
L.~Verdoliva, R.~Gaetano, G.~Ruello, and G.~Poggi, ``Optical-driven nonlocal
  sar despeckling,'' \emph{IEEE Geosci. Remote Sens. Lett.}, vol.~12, no.~2,
  pp. 314--318, Feb. 2015.

\bibitem{JSTAR2018SAROPT}
N.~Merkle, S.~Auer, R.~M¨¹ller, and P.~Reinartz, ``Exploring the potential of
  conditional adversarial networks for optical and sar image matching,''
  \emph{IEEE J. Sel.Topics Appl. Earth Observ. Remote Sens.}, vol.~PP, no.~99,
  pp. 1--10, 2018.

\bibitem{ICML2015BN}
S.~Ioffe and C.~Szegedy, ``Batch normalization: Accelerating deep network
  training by reducing internal covariate shift,'' in \emph{Proceedings of the
  32nd International Conference on Machine Learning}, vol.~37.\hskip 1em plus
  0.5em minus 0.4em\relax PMLR, 07--09 Jul 2015, pp. 448--456.

\bibitem{He2016Resnets}
K.~He, X.~Zhang, S.~Ren, and J.~Sun, ``Deep residual learning for image
  recognition,'' in \emph{IEEE Conference on Computer Vision and Pattern
  Recognition (CVPR)}, July 2016, pp. 770--778.

\bibitem{NIPS2012CNN}
A.~Krizhevsky, I.~Sutskever, and G.~E. Hinton, ``Imagenet classification with
  deep convolutional neural networks,'' in \emph{Advances in Neural Information
  Processing Systems 25}, 2012, pp. 1097--1105.

\bibitem{Isola_2017_CVPR}
P.~Isola, J.-Y. Zhu, T.~Zhou, and A.~A. Efros, ``Image-to-image translation
  with conditional adversarial networks,'' in \emph{The IEEE Conference on
  Computer Vision and Pattern Recognition (CVPR)}, July 2017.

\bibitem{NIPS2014GAN}
I.~Goodfellow, J.~Pouget-Abadie, M.~Mirza, B.~Xu, D.~Warde-Farley, S.~Ozair,
  A.~Courville, and Y.~Bengio, ``Generative adversarial nets,'' in
  \emph{Advances in neural information processing systems}, 2014, pp.
  2672--2680.

\bibitem{ICLR2015DCGAN}
A.~Radford, L.~Metz, and S.~Chintala, ``Unsupervised representation learning
  with deep convolutional generative adversarial networks,'' \emph{arXiv
  preprint arXiv:1511.06434}, 2015.

\bibitem{TIP2017DnCNN}
K.~Zhang, W.~Zuo, Y.~Chen, D.~Meng, and L.~Zhang, ``Beyond a gaussian denoiser:
  Residual learning of deep cnn for image denoising,'' \emph{IEEE Trans. Image
  Process.}, vol.~26, no.~7, pp. 3142--3155, Jun. 2017.

\bibitem{ECCV2016Lifeifei}
J.~Johnson, A.~Alahi, and L.~Fei-Fei, ``Perceptual losses for real-time style
  transfer and super-resolution,'' in \emph{Computer Vision ECCV 2016}, pp.
  694--711.

\bibitem{dlr119293}
P.~Ghamisi and N.~Yokoya, ``Img2dsm: Height simulation from single imagery
  using conditional generative adversarial nets,'' \emph{IEEE Geoscience and
  Remote Sensing Letters}, vol.~PP, no.~99, pp. 1--5, 2018.

\bibitem{zuhlke2015snap}
M.~Zuhlke, N.~Fomferra, C.~Brockmann, M.~Peters, L.~Veci, J.~Malik, and
  P.~Regner, ``Snap (sentinel application platform) and the esa sentinel 3
  toolbox,'' in \emph{Sentinel-3 for Science Workshop}, vol. 734, Dec. 2015,
  p.~21.

\bibitem{He2016TGRS}
W.~He, H.~Zhang, L.~Zhang, and H.~Shen, ``Total-variation-regularized low-rank
  matrix factorization for hyperspectral image restoration,'' \emph{IEEE Trans.
  Geosci. Remote Sens.}, vol.~54, no.~1, pp. 178--188, Jan. 2016.

\end{thebibliography}
\end{document}